\documentclass[10pt,journal,compsoc]{IEEEtran}
\ifCLASSOPTIONcompsoc
  \usepackage[nocompress]{cite}
\else
  \usepackage{cite}
\fi
\ifCLASSINFOpdf
\else
\fi

\usepackage{gensymb}
\usepackage{graphicx}
\usepackage{amsfonts}
\usepackage{amsmath}
\usepackage{tabularx}
\usepackage{multirow}
\usepackage{soul,color}
\usepackage[T1]{fontenc}

\begin{document}
%
\title{Whole-body Detection, Recognition and Identification at Altitude and Range}
%
%
%
%

\author{Siyuan Huang, Ram Prabhakar Kathirvel,
        Chun Pong Lau,
        and~Rama Chellappa,~\IEEEmembership{Life~Fellow,~IEEE}
\IEEEcompsocitemizethanks{\IEEEcompsocthanksitem S. Huang, R. P. Kathirvel and R. Chellappa are with Whiting School of Engineering at Johns Hopkins University, Baltimore,
MD, 21218.\protect\\
C. P. Lau is with the School of Data Science at City University of Hong Kong, Hong Kong. \\
E-mail: \{shuan124, rprabha3, rchella4\}@jhu.edu, cplau27@cityu.edu.hk
}
\thanks{Manuscript received April 19, 2005; revised August 26, 2015.}}

%
%

\markboth{Journal of \LaTeX\ Class Files,~Vol.~14, No.~8, August~2015}%
{Shell \MakeLowercase{\textit{et al.}}: Bare Demo of IEEEtran.cls for Biometrics Council Journals}
%



\IEEEtitleabstractindextext{%
\begin{abstract}
In this paper, we address the challenging task of whole-body biometric detection, recognition, and identification at distances of up to 500m and large pitch angles of up to 50\degree. We propose an end-to-end system evaluated on diverse datasets, including the challenging Biometric Recognition and Identification at Range (BRIAR) dataset. Our approach involves pre-training the detector on common image datasets and fine-tuning it on BRIAR's complex videos and images. After detection, we extract body images and employ a feature extractor for recognition. We conduct thorough evaluations under various conditions, such as different ranges and angles in indoor, outdoor, and aerial scenarios. Our method achieves an average F1 score of 98.29\% at IoU = 0.7 and demonstrates strong performance in recognition accuracy and true acceptance rate at low false acceptance rates compared to existing models. On a test set of 100 subjects with 444 distractors, our model achieves a rank-20 recognition accuracy of 75.13\% and a TAR@1\%FAR of 54.09\%.
\end{abstract}

}

\maketitle

\IEEEdisplaynontitleabstractindextext

%
\IEEEpeerreviewmaketitle

\IEEEraisesectionheading{\section{Introduction}\label{sec:introduction}}

%
%
%
%
\label{sec:intro}
\IEEEPARstart{S}{ince} the introduction of YOLO \cite{redmon2016you}, R-CNN \cite{girshick2014rich}, and ResNet \cite{he2016deep}, object/person detection, and identification based on deep learning has received extensive attention \cite{he2015spatial,girshick2015fast,ren2015faster,liu2016ssd,lin2017feature,lin2017focal}. Similar to general object detection, human body detection identifies the position of specific body object. Additionally, body detection typically has challenges such as occlusion, appearance changes, and background clutter \cite{khan2021deep}. Body detection is useful for various practical applications, including autonomous driving \cite{su2015sparse}, search and rescue \cite{doherty2007uav,rudol2008human}, and verification and recognition \cite{nguyen2017person} based on biometric information. 

\begin{figure}[!t]
  \centering
   \includegraphics[width=\linewidth]{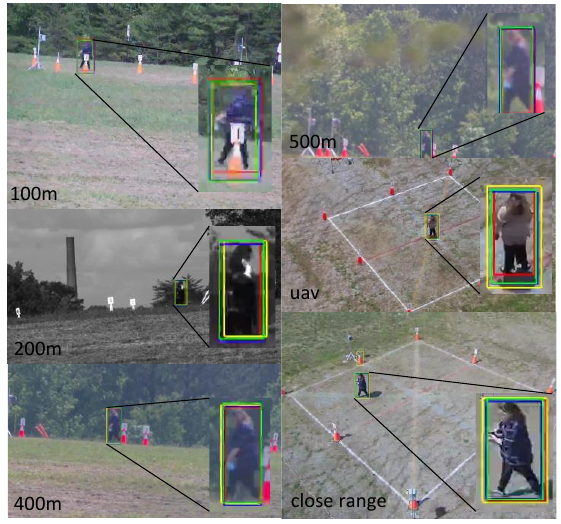}
   \caption{Qualitative results on challenging BRIAR dataset \cite{cornett2022expanding}. Compared with general datasets, BRIAR data has many challenges such as grayscale, blur, occlusion, camera motion, and incomplete human body. Left: original BRIAR datasets at altitudes and ranges. Right: Detection results of our models. Red: ground truth. Other colors: different model detections.}
   \label{fig:1}
\end{figure}

Compared with general object detection datasets (such as COCO \cite{lin2014microsoft}), a key challenge of body detection is processing data from cameras located at different angles, altitudes, locations, and ranges. Fig. \ref{fig:1} shows body images acquired by cameras located at multiple ranges in the field and the corresponding detection results. In some situations, human bodies are even incomplete.

Although there is a vast number of works based on various traditional algorithms \cite{gao2012visual,dalal2006human,dalal2005histograms,bhangale2014human,beiping2011fast,tong2013upper,li2013integrating} to detect bodies. They do not perform well on data collected at range and altitude. Deep learning methods \cite{kim2015human,chen2016kinect,ouyang2015partial,zhang2016faster,li2017pedestrian,du2017fused,wu2017variant} can extract better features and improve detection precision. While deep learning methods have almost monopolized the area of object detection, they are less explored in body detection. For example, among existing methods, only a few \cite{liu2023farsight,guo2023multi} have explored body detection at high altitudes and long ranges.

Motivated by the aforementioned challenges, we propose an end-to-end approach that can detect and identify human bodies at ranges up to 500m and a pitch angle up to 50\degree. Our method requires the model to be pre-trained on a small number of public datasets and then fine-tuned on BRIAR. In both processes, the first stage of the model extracts features and generates proposals. In the second stage, the model detects the body based on these proposals. Finally, the model generates features from body images cropped from the detector. Our main contributions are:
\begin{itemize}
    \item Our method synthesizes the features of public datasets and BRIAR data so that the model can obtain good feature representations without extensive training on large public datasets or BRIAR.
    \item We explore body detection and identification in detail at different altitudes, angles, scenarios, and ranges. The proposed method can detect and identify human bodies under all these challenges with satisfactory accuracy.
    \item Experiments show that our model can maintain an F1 score of $> 0.98$ for BRIAR in almost all cases under different IoU thresholds, as well as a rank-20 accuracy of 90.36\% and a TAR@1\%FAR of 49.26\% on the test set with a large number of distractors, indicating that the model is effective, robust, and stable at different ranges and angles. 
\end{itemize}

\section{Related Work}

\subsection{Body Detection}
HOG \cite{dalal2005histograms} is one of the popular methods for body detection, which used RGB and optical flow to detect human bodies. Many follow-on works \cite{dalal2006human,bhangale2014human,beiping2011fast,seguin2014pose,li2013integrating} are based on HOG to detect bodies. Kamal \textit{et al.} \cite{kamal2016hybrid} detected occluded bodies using background subtraction. Khan \textit{et al.} \cite{khan2016multiple} used templates and 3D modeling to solve the problem of multi-person detection and occlusion. Liu \textit{et al.} \cite{qiang2016hybrid} solved the problem of lighting and low-quality images through cascade head, shoulder detection, and HOG features. Most of these methods did not take into account body movement, whereas each subject in BRIAR would continuously move in difficult scenes or stand still over a short number of frames.

Kim \textit{et al.} \cite{kim2015human} utilized a CNN to directly extract features and classify videos. Ouyang \textit{et al.} \cite{ouyang2015partial} used a multi-layer Restricted Boltzmann Machine (RBM) to extract features and detect bodies at multiple levels, addressing challenges due to occlusion and shadows. Zhang \textit{et al.} \cite{zhang2016faster} used a Faster R-CNN to achieve more accurate and specialized body system. Li \textit{et al.} \cite{li2017pedestrian} replaced the CNN of Faster R-CNN by a dilated CNN to achieve body detection at a lower resolution. Li \textit{et al.} \cite{li2017scale} and Du \textit{et al.} \cite{du2017fused} fused features extracted by multiple R-CNN and SSD, respectively, to detect bodies at arbitrary scales. However, these methods have not been explored in detail for datasets at high altitudes and long ranges.

\begin{figure*}[!t]
  \centering
   \includegraphics[width=\linewidth]{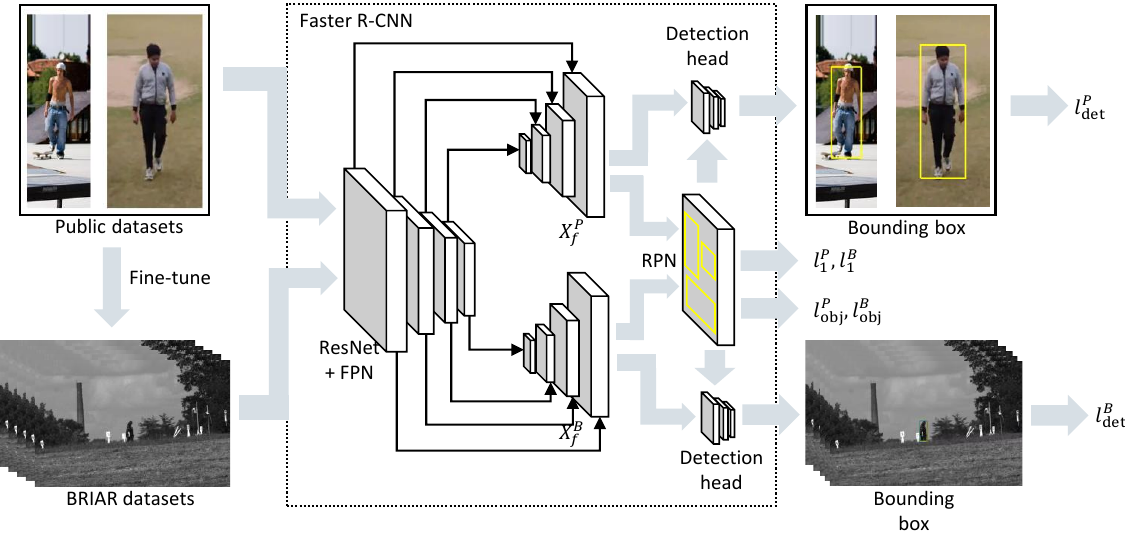}
   \caption{The general architecture of our method involves using annotations of detections from both public datasets and BRIAR to learn a general semantic representation. Once the detector is trained, it is used to generate cropped body images from raw videos, which will then be utilized for recognition and identification.}
   \label{fig:2}
\end{figure*}

\subsection{Recognition and Identification}
Image and video-based methods for person identification and recognition have been developed over the years. Several image-based models \cite{he2020fastreid,luo2019strong,martinel2019aggregating, wieczorek2021unreasonable} have shown impressive results on image-based benchmarks. More complex is the video-based task, which still requires significant research \cite{eom2021video,gu2022clothes,he2021dense,nambiar2019gait,gu2020appearance,hou2020temporal,hou2019vrstc,hou2020iaunet,wang2014person}. However, these methods do not provide any solutions for the complex acquisition conditions mentioned above, and their performance significantly deteriorates at long ranges and high altitudes. While some works attempt to address various difficulties, such as different angles \cite{jin2020uncertainty,sun2019dissecting}, different poses \cite{ge2018fd,qian2018pose,su2017pose}, and occlusions \cite{jin2020semantics,zhang2019densely,he2018deep,miao2019pose,zheng2015partial,zhuo2018occluded}, these methods are not effective when clothing variations are present.

Recently, researchers focusing on human recognition have paid attention to variations in clothing. These efforts mainly fall into two categories: extracting clothing-independent features from RGB images \cite{zhang2019gait,zheng2019joint,huang2019celebrities,huang2019beyond} and using various multimodal information \cite{qian2020long,hong2021fine,jin2022cloth,fan2020learning,yang2019person,chen2021learning} to extract features roust to variations in clothing. In the former, the latest model is Gu \textit{et al.}'s CAL \cite{gu2022clothes}, which used a clothes-based adversarial loss to prevent the clothing classifier from identifying the same person wearing different attire, thereby extracting features that are not related to clothing. In the realm of multi-modality, Jin \textit{et al.} \cite{jin2022cloth} used gait information to identify the body, while Arkushin \textit{et al.} \cite{arkushin2022reface} used face to identify the body. They have all achieved state-of-the-art performance on LTCC at the time of publication. However, they still have not solved the problem of directly identifying persons from raw, complex video data or addressing challenges encountered in real videos, such as height, long distances, and low resolution. Therefore, in our work, we aim to simultaneously consider these challenges to achieve good recognition performance across challenging acquisition scenarios.

\section{Methods}
\label{sec:methods}
The goal of this work is to implement robust detection and recognition systems for whole-body-based biometrics using datasets collected at different altitudes and ranges. Fig. \ref{fig:2} shows the architecture of our method. Specifically, it is based on Faster R-CNN \cite{ren2015faster} and ResNet-50, to learn common features of public datasets and BRIAR. We first train on public datasets to obtain initial body feature information. Given an input $X^P \in \mathbb{R}^{C \times H \times W}$ which is sampled from public datasets, where $C$ is the channel size, $H$ and $W$ are height and width respectively. We extract features $X_f^P$ from it with a feature extractor $f$. $X_f^P$ can be multiple layers when the Feature Pyramid Networks (FPN) \cite{lin2017feature} is used. Then, we use Region Proposal Networks (RPN) \cite{ren2015faster} to calculate proposal $b^P \in \mathbb{R}^4$ and its objectness score $o^P \in \mathbb{R}^2$ for $X_f^P$. $b^P$ and $o^P$ each have a loss. For $b^P$, RPN calculates its smoothed L1 loss \cite{girshick2015fast} as
\begin{equation}
    l_1 = 
    \begin{cases}
    \frac{(b^P - \hat{b}^P)}{2 \beta}, & \text{if } |b^P - \hat{b}^P| < \beta \\
    |b^P - \hat{b}^P| - \frac{\beta}{2}, & \text{otherwise}
    \end{cases}
\end{equation}
where $\hat{b}^P$ is the ground truth bounding box, $\beta$ is smooth threshold. For $o^P$, RPN calculates its binary cross entropy
\begin{equation}
    l_{\text{obj}} = - [\hat{o}^P \log o^P + (1 - \hat{o}^P) \log (1 - o^P)]
\end{equation}
where $\hat{o}^P$ is the label of whether there is an object in the current proposal. Finally, the ROI head calculates the probability that the proposal belongs to a certain class. Since we target the specific class of body, we only have two classes in our approach, person and background. This part is also updated with binary cross entropy loss $l_{\text{det}}$. The final total loss is
\begin{equation}
    \mathcal{L} = l_1 + l_{\text{obj}} + l_{\text{det}}.
\end{equation}

After training on the public dataset, we fine-tune the model on BRIAR. For BRIAR's input $X^B$, the fine-tuning process is the same as was used for the public dataset. We train on multiple public and BRIAR datasets to obtain more synthesized features. However, the sizes of different datasets are quite different. In order to allow the model to sample more evenly, especially focusing on datasets with small sizes, a sampling strategy is employed. In this paper, we adopt the following strategy. Assuming that dataset $d_i$ has $|d_i|$ images or videos, the probability that a certain image or video is sampled from this dataset is $\frac{1}{|d_i|}$. Its final overall sampling probability is normalized on all datasets. Since the size of BRIAR is small at some ranges (e.g., 500m), this is crucial to improving the overall performance on the BRIAR data. Each video is sampled and then a fixed number of frames is uniformly taken to form $X$. All $X$ go through a series of augmentation before fed to the the feature extractor.

\begin{figure*}
    \centering
    \includegraphics[width=1\linewidth]{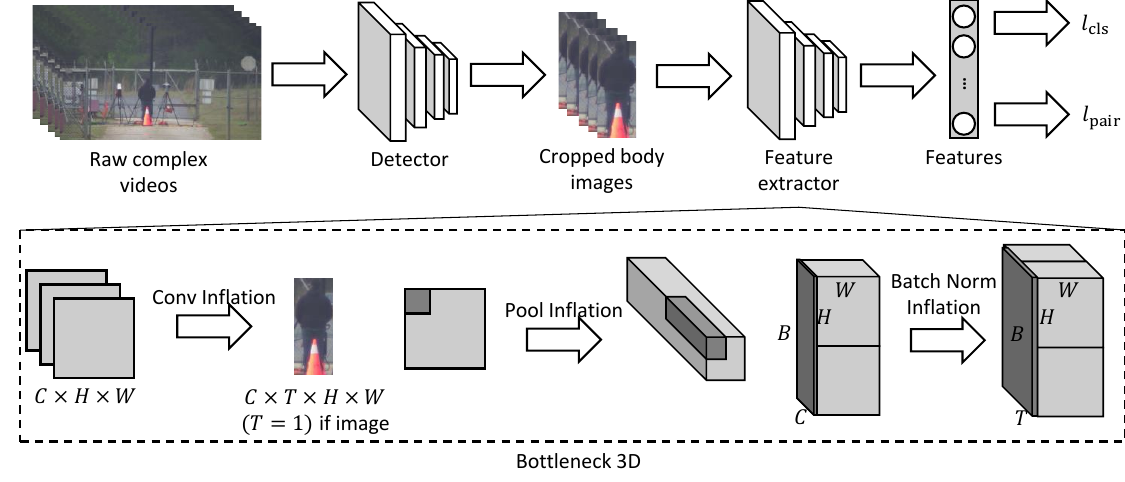}
    \caption{The pipeline of the recognition model. We leverage an inflation to make sure the model can deal with images and videos simultaneously.}
    \label{fig:framework_rec}
\end{figure*}

Fig. \ref{fig:framework_rec} shows the pipeline of the recognition model. The input to the recognition model is $X_{\mathrm{Rec}} \in \mathbb{R}^{B \times C \times T \times H \times W}$, which is the body image sequence cropped from the detector, where $T$ is the length of the time dimension. We adopt a strategy of sampling n people in each batch and $k$ videos for each person, so that $B = nk$. On the other hand, since videos are often longer ($\sim 2k$ to $3k$ frames per video), it is computationally inefficient to directly load all the frames of each video. Furthermore, there is a lot of redundant information in successive frames. Therefore, we take one frame for every fixed stride. Since the length of each video is inconsistent, in order to ensure that all video frames in a batch are of the same length, we also record the index of valid frames of each video. If the length of a video exceeds $T$, we directly take a random frame, and then consecutive $T$ frames are fed to the extractor. Otherwise, we pad 0 to make it $T$. Since each person has controlled images of simple situations (such as images of standing indoors), in addition to complex videos, we also include these images during training. These images are equivalent to a video with a length of 1, so we zero pad them. The extractor ultimately outputs $X_f^{\mathrm{REC}} \in \mathbb{R}^{B \times d}$ features, where $d$ is the feature dimension. Finally, we calculate the $l_{\mathrm{cls}}$ and $l_{\mathrm{pair}}$ of the feature to achieve re-identification, where $l_{\mathrm{cls}}$ is the cross entropy of subjects, and $l_{\mathrm{pair}}$ is the triplet loss \cite{hermans2017defense}. The final loss is
\begin{equation}
    \mathcal{L}_{\text{rec}} = l_{\text{cls}} + l_{\text{pair}}.
\end{equation}

\begin{figure*}[!t]
  \centering
   \includegraphics[width=\linewidth]{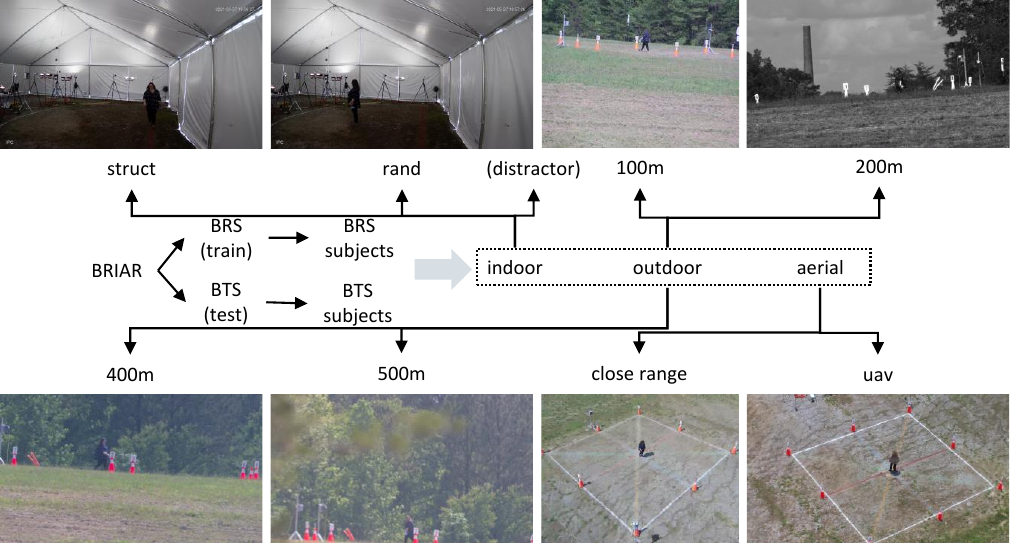}
   \caption{BRIAR's structure tree. Each experimental field has some colored lines. \textit{struct} requires subjects to walk along these lines, while in \textit{rand} subjects walk freely. Note that the data for outdoor environments is the most complex. All subjects are smaller due to acquisition at large range. 200m videos pose challenges due to occlusion. A considerable portion of 500m subjects is at the very bottom or even the corner of frames. Furthermore, large number of outdoor videos suffer from camera shake regardless of ranges, and the shake increases significantly as camera range increases. These issues make robust detection on BRIAR challenging. All images shown have subjects who have given consent.}
   \label{fig:3}
\end{figure*}

\section{Experiments}

\subsection{Datasets}

\textbf{BRIAR.} The BRIAR dataset \cite{cornett2022expanding} consists of images and videos of people moving in various situations, and is divided into face and whole body. The entire dataset has $> 350,000$ images and $1,300$ hours of videos. A total of $~1000$ subjects participated in the production of the entire dataset. Specifically, the BRIAR data can be divided into multiple datasets according to different altitudes, ranges, environments, and actions. The BRIAR data has three environments: indoor, outdoor and aerial (Fig. \ref{fig:3}). The camera in the indoor environment is in a relatively fixed position, and various interferences will be relatively small. Therefore, indoor images and videos are relatively good for training. The indoor subjects have two kinds of movements. First, each subject walks along a preset path on a field. They then walk randomly on the same field. These two constitute two datasets, \textit{struct} and \textit{rand}. In addition to the normally collected data, there is also a dedicated indoor distractor dataset. This dataset is full of images that are processed under various perturbations or scenarios that differ from the training set, which is equivalent to ``physical" augmentation.

Datasets for the outdoor environment are more complex. According to different camera ranges, the outdoor data is divided into four datasets: \textit{100m}, \textit{200m}, \textit{400m}, and \textit{500m}. Since interferences such as outdoor ambient light and occlusion are much stronger in outdoors, these datasets usually challenge the detector. The quality of outdoor videos is significantly lower than indoor videos (Fig. \ref{fig:3}). Therefore, our sampling strategy can focus more on feature learning in difficult outdoor environments. Finally, for aerial data, based on different altitudes and angles, BRIAR data can be divided into \textit{close range} and Unmanned Aerial Vehicles (\textit{UAV}), where close range represents aerial videos near the ground, and UAV represents aerial videos with higher altitudes. Both outdoor and aerial data have struct and rand actions. Fig. \ref{fig:3} shows the structure of BRIAR data as a tree.

The annotation information for BRIAR data includes various attributes of subjects such as birth date, gender, weight, etc. and ground truth bounding boxes. For our experiments, we only need the bounding boxes. We used data from the whole body part. The information on all datasets of BRIAR we use in this paper is shown in Table \ref{tab:1}.

\begin{table}[]
\centering
\begin{tabular}{lcc}
\hline
\textbf{Dataset} & \multicolumn{1}{c}{\textbf{Training}} & \textbf{Test} \\ \hline
struct           & 2,268                                 & 1,225         \\
rand             & 2,257                                 & 1,208         \\
100m             & 729                                   & 382           \\
200m             & 887                                   & 486           \\
400m             & 767                                   & 351           \\
500m             & 695                                   & 407           \\
close range      & 3,553                                 & 1,843         \\
UAV              & 119                                   & 75            \\ \hline
\end{tabular}
\caption{The training and test set size of BRIAR datasets used in this paper.}
\label{tab:1}
\end{table}

\textbf{COCO and Visym.} For public datasets, we mainly use COCO \cite{lin2014microsoft} and Visym \cite{byrne2022fine}. COCO is one of the most classic datasets for object detection tasks, with the advantages of large size and various classes. Visym is a large public dataset dedicated to people, with millions of video clips of hundreds of human activities. Since one of our goals is to reduce training costs while at the same time learn the representations of BRIAR datasets faster, we only extract very few images from COCO and Visym (at least two orders of magnitudes smaller than the full datasets). Specifically, for COCO, we select 567 and 64 images with clear bodies as training and test sets, respectively. For some of the images, we manually re-labeled the accurate bounding box. For Visym, we select 10,430 and 1,064 images from videos of people walking as training and test sets. Section 4.3 will show that with just this few of public information, our fine-tuned model's results on BRIAR are completely comparable to pre-trained models trained on full COCO.

\subsection{Model Settings}

We mainly use the Faster R-CNN \cite{ren2015faster} based on three model settings: pre-trained (PRE), fine-tuned (FT), and from scratch (SCR). Among them, the pre-trained model refers to the model trained only on public datasets we choose, without any BRIAR information. The fine-tuned model refers to fine-tuning on BRIAR after the previous pre-trained model is trained, and the scratched model refers to a model that is directly trained on BRIAR data without public dataset information. During training, all models first sample videos according to the sampling strategy discussed in Sec. \ref{sec:methods}, and then randomly select five frames from each video. At test time, for each video in the test set, we sample frames at fixed time intervals. Therefore, the number of frames taken by each video is different. In order to balance the number of frames taken out for each dataset and speed up the test, time intervals in different datasets are different. For struct, rand, and close range, we take one frame for every 300 frames ($\sim$10s, if FPS $= 30$). For other datasets, models take one frame every 150 frames. Since the BRIAR data is about 1 - 2 minutes per video, at the end about 10 - 20 frames will remain. We use the F1 score as the final evaluation metric. Data augmentations include grayscale ($p = 0.25$), brightness, saturation, contrast, horizontal flip ($p = 0.5$), and RGB shift ($p = 0.3$, limit $= \pm 30$ RGB values).

The feature extractor of our model is based on ResNet-50 \cite{he2016deep} and the 5-layer FPN \cite{lin2017feature}. The $\beta$ of L1 loss is $\frac{1}{9}$. The optimizer is SGD with momentum 0.9. The initial learning rate is 0.00001. There is l2 regularization with a decay rate $=0.0001$. The pre-trained model takes about 11 min/epoch on 8x NVIDIA RTX A5000. Fine-tuned and scratched models take about 3 h/epoch on 7x GeForce RTX 3090 or 6x A5000. The fine-tuned model was finally trained for ten epochs, while the scratched model was trained for twenty-five epochs. Batch size $=1$ was used for each GPU. We also compare performances with the Faster R-CNN implementation of Detectron2 (DET2) \cite{wu2019detectron2} based on complete COCO training. We adopted the model that was closest to the structure of our current model.

The recognition model we use is built using ResNet-50. We use the optimizer Adam with an initial learning rate of $1 \times 10^{-4}$. Additionally, we apply l2 regularization with a decay rate of $5 \times 10^{-4}$. The batch size is 16, which corresponds to 4 people $\times 4$ videos per person. We compare our model with the state-of-the-art CAL.

\begin{table}[t]
\centering
\begin{tabularx}{\linewidth}{X|XX|XX|XX}
\hline
             & \multicolumn{2}{c|}{\textbf{0.35}} & \multicolumn{2}{c|}{\textbf{0.5}} & \multicolumn{2}{c}{\textbf{0.7}} \\
\textbf{model} & \textbf{struct}  & \textbf{rand}  & \textbf{struct}  & \textbf{rand}  & \textbf{struct}  & \textbf{rand} \\ \hline
PRE  & 91.84            & 97.15          & 91.76            & 97.07          & 86.34           & 94.43          \\
FT   & 99.13            & 99.82          & 99.11            & 99.81          & 98.73           & 99.60          \\
SCR & 99.17            & 99.91          & 99.17            & 99.91          & 98.80           & 99.85          \\
DET2   & 98.77            & 99.94          & 98.77            & 99.94          & 98.67           & 99.91          \\ \hline
\end{tabularx}
\caption{F1 score comparison in indoor datasets of BRIAR under different IoU thresholds. Low IoU will help models detect bodies, because now a proposal is more likely to be assigned as a true bounding box. However, it will also increase the False Positive rate because now any proposal is more likely to be assigned a true bounding box. For high IoU, the phenomenon is the opposite.}
\label{tab:2}
\end{table}

\subsection{Results}
\label{sec:formatting}

\begin{table*}[t]
\centering
\begin{tabularx}{\linewidth}{X|XXXX|XXXX|XXXX}
\hline
               & \multicolumn{4}{c|}{\textbf{0.35}}                             & \multicolumn{4}{c|}{\textbf{0.5}}                             & \multicolumn{4}{c}{\textbf{0.7}}                              \\
\textbf{model} & \textbf{100m} & \textbf{200m} & \textbf{400m} & \textbf{500m} & \textbf{100m} & \textbf{200m} & \textbf{400m} & \textbf{500m} & \textbf{100m} & \textbf{200m} & \textbf{400m} & \textbf{500m} \\ \hline
PRE            & 93.19         & 88.53         & 87.01         & 82.09         & 93.02         & 88.41         & 86.77         & 81.97         & 90.55         & 85.06         & 82.57         & 76.57         \\
FT             & 98.90         & 99.74         & 99.15         & 96.78         & 98.71         & 99.69         & 99.10         & 96.49         & 98.05         & 98.70         & 97.90         & 94.53         \\
SCR            & 98.91         & 99.76         & 99.32         & 97.18         & 98.82         & 99.71         & 99.32         & 97.00         & 98.44         & 99.08         & 98.70         & 94.80         \\
DET2           & 99.36         & 99.46         & 99.02         & 98.18         & 99.36         & 99.46         & 99.02         & 98.18         & 99.22         & 99.30         & 98.98         & 98.13         \\ \hline
\end{tabularx}
\caption{F1 score comparison in outdoor datasets of BRIAR under different IoU thresholds.}
\label{tab:3}
\end{table*}

\subsubsection{Detection in Indoor}

We discuss the performance of each model under different altitudes and ranges according to the situation. Table \ref{tab:2} shows the F1 score of all models in indoor struct and rand under different IoU thresholds. Since the indoor situation is the simplest in BRIAR and closest to the situation of public datasets, this comparison can directly reflect the preliminary performance of each model. First, under all IoU thresholds, fine-tuned, scratched, and Detectron2 models achieved almost similar and excellent performance ($>98\%$ of F1 score), while the pre-trained model is not as good as the other three, although it also reached $>90\%$ of F1. Since the pre-trained model has no BRIAR information and only has a small part of public datasets, even some of the simplest indoor cases are not detected. Correspondingly, the Detectron2 trained on the complete COCO achieves similar performance to fine-tuned and scratched models. This basically shows that for datasets with more variations, either specific training for its features, or training on large public datasets to obtain an overall feature distribution, is required.

On the other hand, the performances of the fine-tuned, scratched, and Detectron2 models are very stable and do not change much with IoU. This shows that the bounding boxes predicted by these three models are relatively close to the ground truth. In contrast, the performance of the pre-trained model drops significantly under high IoU thresholds, which shows that its prediction is only a rough estimate, and the body has not been accurately localized. Furthermore, all models perform better on rand than struct. We think this may be because in the rand situation people show multiple angles in front of cameras, while in struct mode view angles are very limited. Such differences may cause models to better recognize rand patterns (see Fig. \ref{fig:4} for comparison).

\begin{table}[!t]
\centering
\begin{tabularx}{\linewidth}{X|XX|XX|XX}
\hline
               & \multicolumn{2}{c|}{\textbf{0.35}}   & \multicolumn{2}{c|}{\textbf{0.5}}   & \multicolumn{2}{c}{\textbf{0.7}}    \\
\textbf{model} & \textbf{close range} & \textbf{UAV} & \textbf{close range} & \textbf{UAV} & \textbf{close range} & \textbf{UAV} \\ \hline
PRE            & 94.75                & 70.45        & 94.69                & 69.48        & 91.81                & 61.14        \\
FT             & 99.39                & 99.51        & 99.39                & 99.43        & 98.73                & 96.99        \\
SCR            & 99.36                & 98.93        & 99.36                & 98.93        & 98.91                & 97.35        \\
DET2           & 99.19                & 98.50        & 99.19                & 98.50        & 98.88                & 98.39        \\ \hline
\end{tabularx}
\caption{F1 score comparison in aerial datasets of BRIAR under different IoU thresholds.}
\label{tab:4}
\end{table}


\subsubsection{Detection in Outdoor}

Table \ref{tab:3} shows the F1 score of all models in different outdoor ranges under different IoU thresholds. First, the three models of fine-tuned, scratched, and Detectron2 have similar, excellent, and stable performance ($>98\%$ of F1) in the cases of 100m, 200m, and 400m, while the performance of the pre-trained model in outdoor is significantly lower than for indoor. This validates the conclusions found in indoor datasets. Second, except for 100m, all models in other ranges show that the performance decreases as the range goes farther. The performance of models at 200m is better than that of 100m. This may be because grayscale helps models to distinguish distinct frames, although it also blends the body and background in few frames. In addition, as the IoU threshold increases, the performance of models at 100m is significantly more stable than that at 400m and 500m, which is in line with the findings between interference and range mentioned above.

When the IoU threshold is not high, the fine-tuned model and scratched model are better than Detectron2 at 200m and 400m. But when the IoU threshold is high, the former is worse than the latter. This shows that models with BRIAR information tend to find the matching pattern first, and then consider localization, while models without BRIAR information behave just the opposite. This difference strongly illustrates the influence of BRIAR information on models about learning body features. Finally, we notice that the fine-tuned model and scratched model have significantly lower performance than Detectron2 at 500m, which further shows that at longer ranges, even if the model has the corresponding feature learning ability, if the body is too small or blurred, it is difficult to provide useful information to the model. However, considering that the fine-tuned model achieves comparable performance with the scratched model and Detectron2 while using only very little public dataset information, we think this still shows that our method is efficient and accurate for various scenarios of BRIAR.

\begin{table}[!t]
\centering
\begin{tabular}{l|c|c|c}
\hline
model & \textbf{0.35} & \textbf{0.5} & \textbf{0.7} \\ \hline
PRE   & 89.57         & 89.46        & 86.10        \\
FT    & 99.00         & 98.94        & 98.11        \\
SCR   & 98.94         & 98.90        & 98.29        \\
DET2  & 99.14         & 99.14        & 98.97        \\ \hline
\end{tabular}
\caption{Average F1 score comparison in all BRIAR datasets under different IoU thresholds.}
\label{tab:5}
\end{table}

\subsubsection{Detection in Aerial Data}

\begin{figure*}[!t]
  \centering
   \includegraphics[width=\linewidth]{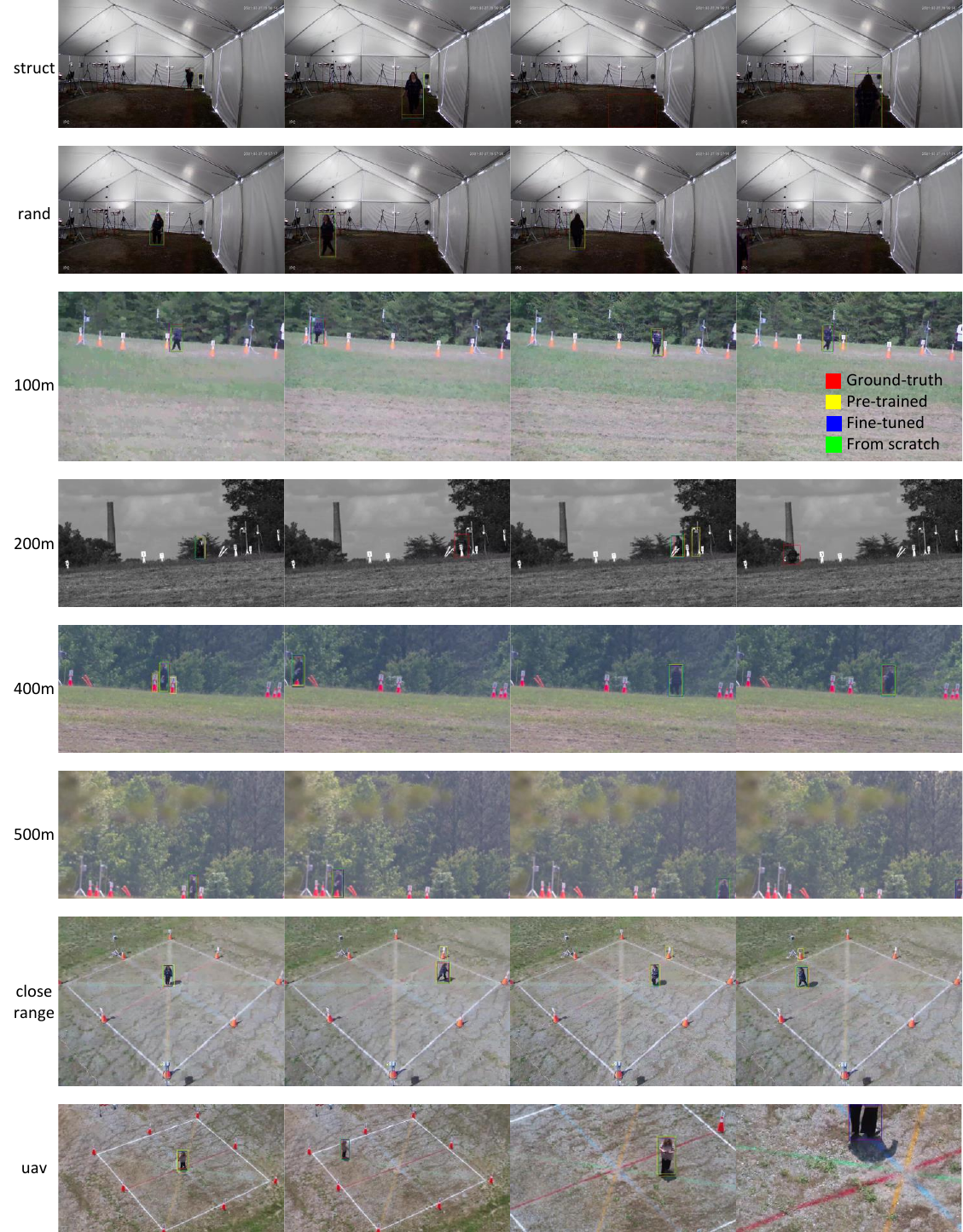}
   \caption{Except for some extreme cases, models with BRIAR information can accurately detect the body on each dataset. This shows that our methods successfully overcome challenges.}
   \label{fig:4}
\end{figure*}

The case of videos collected at altitude is relatively easier. Compared to outdoor results, all models have better performance on aerial video data (table \ref{tab:4}). This may be due to two reasons. First, in aerial datasets, subjects have less occlusions and incomplete bodies (Fig. \ref{fig:3}, Fig. \ref{fig:4}). Second, aerial cameras cover a large area of the ground. Subjects are less likely to fall out of camera range. These reasons make it possible for models to learn features more effectively in aerial situations. It can also be seen from the results at different IoU thresholds that all models are more stable on aerial than outdoor datasets. In summary, on aerial datasets, the fine-tuned model, scratched model and Detectron2 can all achieve $\sim$ 97$+\%$ of the F1 score.

\begin{table*}[!t]
\centering
\begin{tabular}{lcc|cccc|cccc}
\hline
                                          & \textbf{With}     & \textbf{Input} & \multicolumn{4}{c|}{Accuracy}                                           & \multicolumn{4}{c}{TAR@FAR}                                     \\
\textbf{model}                            & \textbf{Face} & \textbf{Type}  & \textbf{Rank-1} & \textbf{Rank-5} & \textbf{Rank-10} & \textbf{Rank-20} & \textbf{0.01\%} & \textbf{0.1\%} & \textbf{1\%} & \textbf{10\%} \\ \hline
DME \cite{guo2023multi}                             & N                 & V              & 24.23           & 51.49           & 64.02            & 74.37            & 0.02          & 0.02         & 2.15       & 22.97        \\
CAL \cite{gu2022clothes}                              & N                 & I              & 20.92           & 42.34           & 54.45            & 66.78            & 0.03          & 0.70         & 5.13       & 28.28        \\ \hline
\multirow{3}{*}{\textbf{BRIARNet}} & N                 & I              & 36.66& 66.86& 78.98& 87.59& 2.79& 11.13& 35.69& 71.28\\
                                          & N                 & I + V          & 29.85           & 58.83           & 68.07            & 77.65            & 1.85          & 8.32         & 32.54       & 73.10        \\
                                          & Y                 & I + V          & \textbf{46.67}           & \textbf{73.75}           & \textbf{82.51}            & \textbf{90.36}            & \textbf{4.13}          & \textbf{15.97}         & \textbf{49.26}       & \textbf{86.78}        \\ \hline

\end{tabular}
\caption{A performance comparison of our model (\textbf{BRIARNet}) with state-of-the-art on Protocol 1, based on the accuracy of different ranks and the True Accept Rate (TAR) at different False Accept Rates (FAR). Protocol 1 has 85 subjects an 100 distractors. Since BRIAR has both face and whole body data, the ``With Face'' item means whether the model is trained with face data. This is not same to the difference between FaceIncluded and FaceRestricted in \cite{liu2023farsight}. All results in this paper is FaceIncluded. The ``Input Type'' is of whether the model uses images and videos.}
\label{tab:6}
\end{table*}

\begin{table*}[!t]
\centering
\begin{tabular}{lcc|cccc|cccc}
\hline
                                          & \textbf{With}     & \textbf{Input} & \multicolumn{4}{c|}{Accuracy}                                           & \multicolumn{4}{c}{TAR@FAR}                                     \\
\textbf{model}                            & \textbf{Face} & \textbf{Type}  & \textbf{Rank-1} & \textbf{Rank-5} & \textbf{Rank-10} & \textbf{Rank-20} & \textbf{0.01\%} & \textbf{0.1\%} & \textbf{1\%} & \textbf{10\%} \\ \hline
\multirow{3}{*}{\textbf{BRIARNet}} & N                 & I              & 12.18  & 24.27  & 25.72   & 37.17   & 2.24      & 7.94      & 22.55      & 43.64      \\
                                          & N                 & I + V          & 11.84  & 24.42  & 30.95   & 38.28   & 1.75 & 8.45 & 22.73 & 48.03 \\
                                          & Y                 & I + V          & \textbf{34.84}  & \textbf{55.93}  & \textbf{64.70}   & \textbf{73.30}   & \textbf{6.61} & \textbf{24.90} & \textbf{53.77} & \textbf{83.03} \\ \hline

\end{tabular}
\caption{A performance comparison of our model on Protocol 2. Protocol 2 presents a more challenging task than Protocol 1 due to its 100 subjects and 444 distractors. Nevertheless, our model performed well, demonstrating its effectiveness in difficult scenarios.}
\label{tab:7}
\end{table*}

We finally show the average results of all models in Table \ref{tab:5}. Note that the average results here are not a simple average of all previous results, but the F1 score of models tested on all BRIAR datasets. It can be seen that the fine-tuned model, scratched model and Detectron2 basically have similar performances. This suggests that our method does not require pre-training on massive public datasets, nor training on briar for a long time from scratch, and can achieve an average of 99\% of F1 on BRIAR datasets. Features learned by the fine-tuned model are sufficient to support downstream applications (such as verification, recognition, etc.).

\subsection{Visualizations of Detection Results}

Fig. \ref{fig:4} shows the detection results of different models in consecutive video frames for each dataset. For many datasets, the pre-trained model has produced a large number of false negatives (such as the fan of images 1-2 in struct, the cone of images 2-4 in close range) and failures (such as in image 2 in 100m it did not detect the body). The fine-tuned model and the scratched model are more performant. However, for many corner cases, a degradation in performance is observed. For example, the subject image 3 in struct has only hairs on the back of her head and blends into the background rather well. At this time, except for the ground truth (red), the naked eye may not be able to accurately determine where the hair region is. Nevertheless, apart from these corner cases, it can be seen that in most indoor, outdoor and aerial scenarios, both the fine-tuned model and the scratched model can detect bodies well. This illustrates the effectiveness of our method.

Models that learn BRIAR information can produce extremely accurate detections in some cases. For example, for image 4 in 500m, the entire subject only occupied dozens of pixels in the corner of the image. For image 4 in UAV, the subject has only legs visible. However, the fine-tuned model can still accurately find the body in these cases, and the difference from ground truth is very small. In summary, using both the generality of the public dataset and the specificity of BRIAR, models can be trained quickly on both sides and learn a more robust representation. This may be very helpful to address the domain shift problem of representation.

\begin{table}[!t]
\centering
\begin{tabularx}{0.5\textwidth}{lXXX|cc}
\hline
\textbf{model}                     & \textbf{With Face} & \textbf{Input Type} & \textbf{BRS} & \textbf{Rank-20} & \textbf{1\% FAR} \\ \hline
CAL \cite{gu2022clothes}                                & Y                  & I + V                   & 1, 2                & 71.18                & 51.87                 \\
FarSight \cite{liu2023farsight}                    & Y                  & I + V               & 1, 2                & 72.91                & 54.00                 \\ \hline
\multirow{4}{*}{\textbf{BRIARNet}} & N                  & I                   & 1, 2                & 37.17                & 22.55                 \\
                                   & N                  & I + V               & 1, 2                & 38.28                & 22.73                 \\
                                   & Y                  & I + V               & 1, 2                & \textbf{73.30}       & 53.77                 \\
                                   & Y                  & I + V               & 1, 2, 3               & \textbf{75.13}       & \textbf{54.09}        \\ \hline
\end{tabularx}
\caption{A performance comparison of our models with state-of-the-art on Protocol 2. The ``BRS'' item is related to which dataset the model uses in training, i.e., different training subsets defined in \cite{cornett2022expanding}.}
\label{tab:8}
\end{table}


\subsection{Recognition and Identification performance}

Table \ref{tab:6} presents a performance comparison of our model with DME \cite{guo2023multi} and CAL \cite{gu2022clothes} on protocol 1. The primary metrics used for comparison are the accuracy of different ranks and the TAR under different FARs. Protocol 1 consists of a total of 185 subjects, including 100 distractors. The key differences among our models are as follows. Model 2 has a larger gallery size and more difficult gallery than model 1, as it has more controlled images. Model 3, on the other hand, is trained on more challenging situations, including 270m, 370m, 490m, 600m, 800m, and 1000m. Additionally, model 3 is trained on face data simultaneously. Although face data may only include a partial body or even only a head, we included it in the training. It is evident that regardless of the model version, our models have achieved substantial performance improvements compared to CAL. Table \ref{tab:7} presents a comparison of our results on the more challenging Protocol 2. This protocol consists of a total of 544 subjects, 444 of which are distractors. Despite the large number of distractors, our model achieved an accuracy of 73.30\% at rank-20 and a TAR@1\%FAR of 53.77\%, demonstrating the effectiveness of our method for complex situations. Furthermore, model 2 outperforms model 1, while model 3 has achieved the highest performance, demonstrating the robustness of our model. 

\begin{figure*}[h]
    \centering
    \includegraphics[width=1\linewidth]{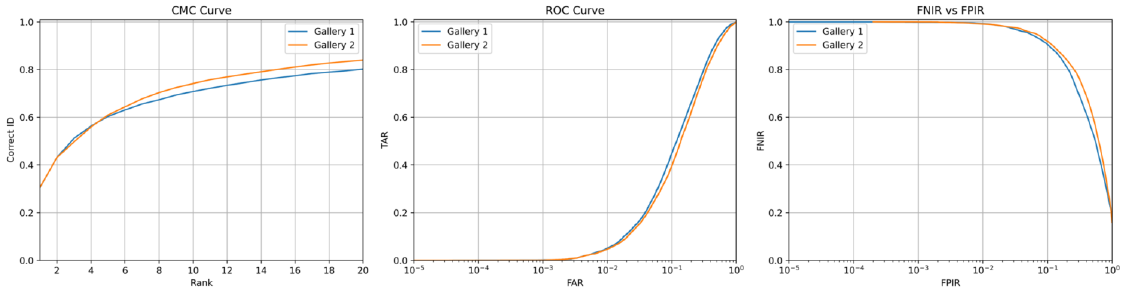}
    \caption{The performance curves of BRIARNet for both galleries on protocol 3.1. BRIARNet have $>= 80\%$ rank-20 accuracy, which is even higher than protocol 2's results. Besides, we also provide results of FNIR vs FPIR here as reference for other works in open-set recognition setting. These results show the robustness of BRIARNet.}
    \label{fig:5}
\end{figure*}

Table \ref{tab:8} summarizes the performance comparison between state-of-the-art models and our models. Since FarSight \cite{liu2023farsight} is a synthesized model of body, gait, and face, we only compare with their pure body version for a fair comparison. Under same setting, our model has better performance than state-of-the-art methods. An interesting finding is that FarSight and BRIARNet have very similar and competitive performance. Considering that the backbone of both models is ResNet-50, this suggests that backbones decide the lower bound of the representation ability of bodies. On the other hand, BRIARNet has better rank-20 accuracy, while FarSight is slightly better in TAR. This suggests that different strategies focus on improving different aspects of representations, thereby moving to different upper bounds. A pure RGB feature representation (BRIARNet) tends to cluster similar identities, while 3D modeling (3DInvarReID \cite{liu2023learning} in FarSight) tends to recognize same identity. Besides, adding more data helps improve all metrics. Compared with our models from Table \ref{tab:6} to \ref{tab:8}, models trained with face images, videos, and various scenarios are better than models trained without them. Among them, face information is essential for good ReID.

To our knowledge, there has been no comprehensive report on whole-body recognition results for Protocol 3.1. In response, we present an evaluation of BRIARNet's performance on Protocol 3.1. Protocol 3.1 comprises two galleries: Gallery 1 consists of 134 subjects and 351 distractors, while Gallery 2 contains 130 subjects and 351 distractors. In Fig. \ref{fig:5}, we display the performance curves of BRIARNet for both galleries. The CMC curve corresponds to rank-$k$ accuracy, the ROC curve represents the TAR vs FAR, and FNIR denotes the fraction of failed mate searches that exceed a predefined threshold \cite{grother2019face}, serving as a metric for assessing open-set recognition performance. It is evident that BRIARNet's performance on the CMC curve remains largely unaffected, and in some cases, it even improves. Furthermore, even when confronted with a substantial number of distractors, BRIARNet maintains a reasonable level of precision (TAR $> 0$ when FAR $> 10^{-3}$) and demonstrates commendable open-set recognition capabilities (FNIR $< 1$ when FPIR $> 10^{-2}$). This underscores the robustness of BRIARNet.

\section{Conclusion}
In this paper, we have successfully performed body detection and identification on real-world datasets, including BRIAR data, under various altitudes and ranges. Our method generates features from public datasets and BRIAR, allowing our models to achieve high performance without exhaustive training on large datasets. By pre-training on a small number of public datasets and fine-tuning on BRIAR, our models can achieve 98\% F1 score within 10 epochs. Additionally, our model achieved an accuracy of 75.13\% and a TAR@1\%FAR of 54.09\%, outperforming state-of-the-art recognition and identification models. Experimental results demonstrate that models obtained through fine-tuning can maintain robust performance under different altitudes, ranges, environments, and actions.


%



\ifCLASSOPTIONcompsoc
  \section*{Acknowledgments}
\else
  \section*{Acknowledgment}
\fi

This research is based upon work supported in part by the Office of the Director of National Intelligence (ODNI),
Intelligence Advanced Research Projects Activity (IARPA), via [2022-21102100005]. The views and conclusions contained herein are those of the authors and should not be interpreted as necessarily representing the official policies, either expressed or implied, of ODNI, IARPA, or the U. S. Government. The US. Government is authorized to reproduce and distribute reprints for governmental purposes notwithstanding any copyright annotation therein.

\ifCLASSOPTIONcaptionsoff
  \newpage
\fi



\bibliographystyle{IEEEtran}
\bibliography{ref}

\begin{thebibliography}{10}
\providecommand{\url}[1]{#1}
\csname url@samestyle\endcsname
\providecommand{\newblock}{\relax}
\providecommand{\bibinfo}[2]{#2}
\providecommand{\BIBentrySTDinterwordspacing}{\spaceskip=0pt\relax}
\providecommand{\BIBentryALTinterwordstretchfactor}{4}
\providecommand{\BIBentryALTinterwordspacing}{\spaceskip=\fontdimen2\font plus
\BIBentryALTinterwordstretchfactor\fontdimen3\font minus \fontdimen4\font\relax}
\providecommand{\BIBforeignlanguage}[2]{{%
\expandafter\ifx\csname l@#1\endcsname\relax
\typeout{** WARNING: IEEEtran.bst: No hyphenation pattern has been}%
\typeout{** loaded for the language `#1'. Using the pattern for}%
\typeout{** the default language instead.}%
\else
\language=\csname l@#1\endcsname
\fi
#2}}
\providecommand{\BIBdecl}{\relax}
\BIBdecl

\bibitem{redmon2016you}
J.~Redmon, S.~Divvala, R.~Girshick, and A.~Farhadi, ``You only look once: Unified, real-time object detection,'' in \emph{Proceedings of the IEEE conference on computer vision and pattern recognition}, 2016, pp. 779--788.

\bibitem{girshick2014rich}
R.~Girshick, J.~Donahue, T.~Darrell, and J.~Malik, ``Rich feature hierarchies for accurate object detection and semantic segmentation,'' in \emph{Proceedings of the IEEE conference on computer vision and pattern recognition}, 2014, pp. 580--587.

\bibitem{he2016deep}
K.~He, X.~Zhang, S.~Ren, and J.~Sun, ``Deep residual learning for image recognition,'' in \emph{Proceedings of the IEEE conference on computer vision and pattern recognition}, 2016, pp. 770--778.

\bibitem{he2015spatial}
------, ``Spatial pyramid pooling in deep convolutional networks for visual recognition,'' \emph{IEEE transactions on pattern analysis and machine intelligence}, vol.~37, no.~9, pp. 1904--1916, 2015.

\bibitem{girshick2015fast}
R.~Girshick, ``Fast r-cnn,'' in \emph{Proceedings of the IEEE international conference on computer vision}, 2015, pp. 1440--1448.

\bibitem{ren2015faster}
S.~Ren, K.~He, R.~Girshick, and J.~Sun, ``Faster r-cnn: Towards real-time object detection with region proposal networks,'' \emph{Advances in neural information processing systems}, vol.~28, 2015.

\bibitem{liu2016ssd}
W.~Liu, D.~Anguelov, D.~Erhan, C.~Szegedy, S.~Reed, C.-Y. Fu, and A.~C. Berg, ``Ssd: Single shot multibox detector,'' in \emph{European conference on computer vision}.\hskip 1em plus 0.5em minus 0.4em\relax Springer, 2016, pp. 21--37.

\bibitem{lin2017feature}
T.-Y. Lin, P.~Doll{\'a}r, R.~Girshick, K.~He, B.~Hariharan, and S.~Belongie, ``Feature pyramid networks for object detection,'' in \emph{Proceedings of the IEEE conference on computer vision and pattern recognition}, 2017, pp. 2117--2125.

\bibitem{lin2017focal}
T.-Y. Lin, P.~Goyal, R.~Girshick, K.~He, and P.~Doll{\'a}r, ``Focal loss for dense object detection,'' in \emph{Proceedings of the IEEE international conference on computer vision}, 2017, pp. 2980--2988.

\bibitem{khan2021deep}
M.~A. Khan, M.~Mittal, L.~M. Goyal, and S.~Roy, ``A deep survey on supervised learning based human detection and activity classification methods,'' \emph{Multimedia Tools and Applications}, vol.~80, no.~18, pp. 27\,867--27\,923, 2021.

\bibitem{su2015sparse}
S.-Z. Su, Z.-H. Liu, S.-P. Xu, S.-Z. Li, and R.~Ji, ``Sparse auto-encoder based feature learning for human body detection in depth image,'' \emph{Signal Processing}, vol. 112, pp. 43--52, 2015.

\bibitem{doherty2007uav}
P.~Doherty and P.~Rudol, ``A uav search and rescue scenario with human body detection and geolocalization,'' in \emph{Australasian Joint Conference on Artificial Intelligence}.\hskip 1em plus 0.5em minus 0.4em\relax Springer, 2007, pp. 1--13.

\bibitem{rudol2008human}
P.~Rudol and P.~Doherty, ``Human body detection and geolocalization for uav search and rescue missions using color and thermal imagery,'' in \emph{2008 IEEE aerospace conference}.\hskip 1em plus 0.5em minus 0.4em\relax Ieee, 2008, pp. 1--8.

\bibitem{nguyen2017person}
D.~T. Nguyen, H.~G. Hong, K.~W. Kim, and K.~R. Park, ``Person recognition system based on a combination of body images from visible light and thermal cameras,'' \emph{Sensors}, vol.~17, no.~3, p. 605, 2017.

\bibitem{lin2014microsoft}
T.-Y. Lin, M.~Maire, S.~Belongie, J.~Hays, P.~Perona, D.~Ramanan, P.~Doll{\'a}r, and C.~L. Zitnick, ``Microsoft coco: Common objects in context,'' in \emph{European conference on computer vision}.\hskip 1em plus 0.5em minus 0.4em\relax Springer, 2014, pp. 740--755.

\bibitem{cornett2022expanding}
D.~Cornett~III, J.~Brogan, N.~Barber, D.~Aykac, S.~Baird, N.~Burchfield, C.~Dukes, A.~Duncan, R.~Ferrell, J.~Goddard \emph{et~al.}, ``Expanding accurate person recognition to new altitudes and ranges: The briar dataset,'' \emph{arXiv preprint arXiv:2211.01917}, 2022.

\bibitem{gao2012visual}
Y.~Gao, M.~Wang, Z.-J. Zha, J.~Shen, X.~Li, and X.~Wu, ``Visual-textual joint relevance learning for tag-based social image search,'' \emph{IEEE Transactions on Image Processing}, vol.~22, no.~1, pp. 363--376, 2012.

\bibitem{dalal2006human}
N.~Dalal, B.~Triggs, and C.~Schmid, ``Human detection using oriented histograms of flow and appearance,'' in \emph{European conference on computer vision}.\hskip 1em plus 0.5em minus 0.4em\relax Springer, 2006, pp. 428--441.

\bibitem{dalal2005histograms}
N.~Dalal and B.~Triggs, ``Histograms of oriented gradients for human detection,'' in \emph{2005 IEEE computer society conference on computer vision and pattern recognition (CVPR'05)}, vol.~1.\hskip 1em plus 0.5em minus 0.4em\relax Ieee, 2005, pp. 886--893.

\bibitem{bhangale2014human}
K.~B. Bhangale and R.~Shekokar, ``Human body detection in static images using hog \& piecewise linear svm,'' \emph{Int. J. Innov. Res. Dev}, vol.~3, no.~6, 2014.

\bibitem{beiping2011fast}
H.~Beiping and Z.~Wen, ``Fast human detection using motion detection and histogram of oriented gradients.'' \emph{J. Comput.}, vol.~6, no.~8, pp. 1597--1604, 2011.

\bibitem{tong2013upper}
R.~Tong, D.~Xie, and M.~Tang, ``Upper body human detection and segmentation in low contrast video,'' \emph{IEEE transactions on circuits and systems for video technology}, vol.~23, no.~9, pp. 1502--1509, 2013.

\bibitem{li2013integrating}
D.~Li, L.~Xu, E.~D. Goodman, Y.~Xu, and Y.~Wu, ``Integrating a statistical background-foreground extraction algorithm and svm classifier for pedestrian detection and tracking,'' \emph{Integrated Computer-Aided Engineering}, vol.~20, no.~3, pp. 201--216, 2013.

\bibitem{kim2015human}
Y.~Kim and T.~Moon, ``Human detection and activity classification based on micro-doppler signatures using deep convolutional neural networks,'' \emph{IEEE geoscience and remote sensing letters}, vol.~13, no.~1, pp. 8--12, 2015.

\bibitem{chen2016kinect}
X.~Chen, K.~Henrickson, and Y.~Wang, ``Kinect-based pedestrian detection for crowded scenes,'' \emph{Computer-Aided Civil and Infrastructure Engineering}, vol.~31, no.~3, pp. 229--240, 2016.

\bibitem{ouyang2015partial}
W.~Ouyang, X.~Zeng, and X.~Wang, ``Partial occlusion handling in pedestrian detection with a deep model,'' \emph{IEEE Transactions on Circuits and Systems for Video Technology}, vol.~26, no.~11, pp. 2123--2137, 2015.

\bibitem{zhang2016faster}
L.~Zhang, L.~Lin, X.~Liang, and K.~He, ``Is faster r-cnn doing well for pedestrian detection?'' in \emph{European conference on computer vision}.\hskip 1em plus 0.5em minus 0.4em\relax Springer, 2016, pp. 443--457.

\bibitem{li2017pedestrian}
J.~Li, Y.~Wu, J.~Zhao, L.~Guan, C.~Ye, and T.~Yang, ``Pedestrian detection with dilated convolution, region proposal network and boosted decision trees,'' in \emph{2017 International Joint Conference on Neural Networks (IJCNN)}.\hskip 1em plus 0.5em minus 0.4em\relax IEEE, 2017, pp. 4052--4057.

\bibitem{du2017fused}
X.~Du, M.~El-Khamy, J.~Lee, and L.~Davis, ``Fused dnn: A deep neural network fusion approach to fast and robust pedestrian detection,'' in \emph{2017 IEEE winter conference on applications of computer vision (WACV)}.\hskip 1em plus 0.5em minus 0.4em\relax IEEE, 2017, pp. 953--961.

\bibitem{wu2017variant}
S.~Wu, H.-S. Wong, and S.~Wang, ``Variant semiboost for improving human detection in application scenes,'' \emph{IEEE Transactions on Circuits and Systems for Video Technology}, vol.~28, no.~7, pp. 1595--1608, 2017.

\bibitem{liu2023farsight}
F.~Liu, R.~Ashbaugh, N.~Chimitt, N.~Hassan, A.~Hassani, A.~Jaiswal, M.~Kim, Z.~Mao, C.~Perry, Z.~Ren \emph{et~al.}, ``Farsight: A physics-driven whole-body biometric system at large distance and altitude,'' \emph{arXiv preprint arXiv:2306.17206}, 2023.

\bibitem{guo2023multi}
Y.~Guo, C.~Peng, C.~P. Lau, and R.~Chellappa, ``Multi-modal human authentication using silhouettes, gait and rgb,'' in \emph{2023 IEEE 17th International Conference on Automatic Face and Gesture Recognition (FG)}.\hskip 1em plus 0.5em minus 0.4em\relax IEEE, 2023, pp. 1--7.

\bibitem{seguin2014pose}
G.~Seguin, K.~Alahari, J.~Sivic, and I.~Laptev, ``Pose estimation and segmentation of multiple people in stereoscopic movies,'' \emph{IEEE transactions on pattern analysis and machine intelligence}, vol.~37, no.~8, pp. 1643--1655, 2014.

\bibitem{kamal2016hybrid}
S.~Kamal and A.~Jalal, ``A hybrid feature extraction approach for human detection, tracking and activity recognition using depth sensors,'' \emph{Arabian Journal for science and engineering}, vol.~41, no.~3, pp. 1043--1051, 2016.

\bibitem{khan2016multiple}
M.~H. Khan, K.~Shirahama, M.~S. Farid, and M.~Grzegorzek, ``Multiple human detection in depth images,'' in \emph{2016 IEEE 18th International Workshop on Multimedia Signal Processing (MMSP)}.\hskip 1em plus 0.5em minus 0.4em\relax IEEE, 2016, pp. 1--6.

\bibitem{qiang2016hybrid}
L.~Qiang, W.~Zhang, L.~Hongliang, and K.~N. Ngan, ``Hybrid human detection and recognition in surveillance,'' \emph{Neurocomputing}, vol. 194, pp. 10--23, 2016.

\bibitem{li2017scale}
J.~Li, X.~Liang, S.~Shen, T.~Xu, J.~Feng, and S.~Yan, ``Scale-aware fast r-cnn for pedestrian detection,'' \emph{IEEE transactions on Multimedia}, vol.~20, no.~4, pp. 985--996, 2017.

\bibitem{he2020fastreid}
L.~He, X.~Liao, W.~Liu, X.~Liu, P.~Cheng, and T.~Mei, ``Fastreid: A pytorch toolbox for general instance re-identification,'' \emph{arXiv preprint arXiv:2006.02631}, 2020.

\bibitem{luo2019strong}
H.~Luo, W.~Jiang, Y.~Gu, F.~Liu, X.~Liao, S.~Lai, and J.~Gu, ``A strong baseline and batch normalization neck for deep person re-identification,'' \emph{IEEE Transactions on Multimedia}, vol.~22, no.~10, pp. 2597--2609, 2019.

\bibitem{martinel2019aggregating}
N.~Martinel, G.~Luca~Foresti, and C.~Micheloni, ``Aggregating deep pyramidal representations for person re-identification,'' in \emph{Proceedings of the IEEE/CVF Conference on Computer Vision and Pattern Recognition Workshops}, 2019, pp. 0--0.

\bibitem{wieczorek2021unreasonable}
M.~Wieczorek, B.~Rychalska, and J.~D{\k{a}}browski, ``On the unreasonable effectiveness of centroids in image retrieval,'' in \emph{Neural Information Processing: 28th International Conference, ICONIP 2021, Sanur, Bali, Indonesia, December 8--12, 2021, Proceedings, Part IV 28}.\hskip 1em plus 0.5em minus 0.4em\relax Springer, 2021, pp. 212--223.

\bibitem{eom2021video}
C.~Eom, G.~Lee, J.~Lee, and B.~Ham, ``Video-based person re-identification with spatial and temporal memory networks,'' in \emph{Proceedings of the IEEE/CVF International Conference on Computer Vision}, 2021, pp. 12\,036--12\,045.

\bibitem{gu2022clothes}
X.~Gu, H.~Chang, B.~Ma, S.~Bai, S.~Shan, and X.~Chen, ``Clothes-changing person re-identification with rgb modality only,'' in \emph{Proceedings of the IEEE/CVF Conference on Computer Vision and Pattern Recognition}, 2022, pp. 1060--1069.

\bibitem{he2021dense}
T.~He, X.~Jin, X.~Shen, J.~Huang, Z.~Chen, and X.-S. Hua, ``Dense interaction learning for video-based person re-identification,'' in \emph{Proceedings of the IEEE/CVF International Conference on Computer Vision}, 2021, pp. 1490--1501.

\bibitem{nambiar2019gait}
A.~Nambiar, A.~Bernardino, and J.~C. Nascimento, ``Gait-based person re-identification: A survey,'' \emph{ACM Computing Surveys (CSUR)}, vol.~52, no.~2, pp. 1--34, 2019.

\bibitem{gu2020appearance}
X.~Gu, H.~Chang, B.~Ma, H.~Zhang, and X.~Chen, ``Appearance-preserving 3d convolution for video-based person re-identification,'' in \emph{Computer Vision--ECCV 2020: 16th European Conference, Glasgow, UK, August 23--28, 2020, Proceedings, Part II 16}.\hskip 1em plus 0.5em minus 0.4em\relax Springer, 2020, pp. 228--243.

\bibitem{hou2020temporal}
R.~Hou, H.~Chang, B.~Ma, S.~Shan, and X.~Chen, ``Temporal complementary learning for video person re-identification,'' in \emph{Computer Vision--ECCV 2020: 16th European Conference, Glasgow, UK, August 23--28, 2020, Proceedings, Part XXV 16}.\hskip 1em plus 0.5em minus 0.4em\relax Springer, 2020, pp. 388--405.

\bibitem{hou2019vrstc}
R.~Hou, B.~Ma, H.~Chang, X.~Gu, S.~Shan, and X.~Chen, ``Vrstc: Occlusion-free video person re-identification,'' in \emph{Proceedings of the IEEE/CVF conference on computer vision and pattern recognition}, 2019, pp. 7183--7192.

\bibitem{hou2020iaunet}
------, ``Iaunet: Global context-aware feature learning for person reidentification,'' \emph{IEEE Transactions on Neural Networks and Learning Systems}, vol.~32, no.~10, pp. 4460--4474, 2020.

\bibitem{wang2014person}
T.~Wang, S.~Gong, X.~Zhu, and S.~Wang, ``Person re-identification by video ranking,'' in \emph{Computer Vision--ECCV 2014: 13th European Conference, Zurich, Switzerland, September 6-12, 2014, Proceedings, Part IV 13}.\hskip 1em plus 0.5em minus 0.4em\relax Springer, 2014, pp. 688--703.

\bibitem{jin2020uncertainty}
X.~Jin, C.~Lan, W.~Zeng, and Z.~Chen, ``Uncertainty-aware multi-shot knowledge distillation for image-based object re-identification,'' in \emph{Proceedings of the AAAI Conference on Artificial Intelligence}, vol.~34, no.~07, 2020, pp. 11\,165--11\,172.

\bibitem{sun2019dissecting}
X.~Sun and L.~Zheng, ``Dissecting person re-identification from the viewpoint of viewpoint,'' in \emph{Proceedings of the IEEE/CVF conference on computer vision and pattern recognition}, 2019, pp. 608--617.

\bibitem{ge2018fd}
Y.~Ge, Z.~Li, H.~Zhao, G.~Yin, S.~Yi, X.~Wang \emph{et~al.}, ``Fd-gan: Pose-guided feature distilling gan for robust person re-identification,'' \emph{Advances in neural information processing systems}, vol.~31, 2018.

\bibitem{qian2018pose}
X.~Qian, Y.~Fu, T.~Xiang, W.~Wang, J.~Qiu, Y.~Wu, Y.-G. Jiang, and X.~Xue, ``Pose-normalized image generation for person re-identification,'' in \emph{Proceedings of the European conference on computer vision (ECCV)}, 2018, pp. 650--667.

\bibitem{su2017pose}
C.~Su, J.~Li, S.~Zhang, J.~Xing, W.~Gao, and Q.~Tian, ``Pose-driven deep convolutional model for person re-identification,'' in \emph{Proceedings of the IEEE international conference on computer vision}, 2017, pp. 3960--3969.

\bibitem{jin2020semantics}
X.~Jin, C.~Lan, W.~Zeng, G.~Wei, and Z.~Chen, ``Semantics-aligned representation learning for person re-identification,'' in \emph{Proceedings of the AAAI Conference on Artificial Intelligence}, vol.~34, no.~07, 2020, pp. 11\,173--11\,180.

\bibitem{zhang2019densely}
Z.~Zhang, C.~Lan, W.~Zeng, and Z.~Chen, ``Densely semantically aligned person re-identification,'' in \emph{Proceedings of the IEEE/CVF conference on computer vision and pattern recognition}, 2019, pp. 667--676.

\bibitem{he2018deep}
L.~He, J.~Liang, H.~Li, and Z.~Sun, ``Deep spatial feature reconstruction for partial person re-identification: Alignment-free approach,'' in \emph{Proceedings of the IEEE conference on computer vision and pattern recognition}, 2018, pp. 7073--7082.

\bibitem{miao2019pose}
J.~Miao, Y.~Wu, P.~Liu, Y.~Ding, and Y.~Yang, ``Pose-guided feature alignment for occluded person re-identification,'' in \emph{Proceedings of the IEEE/CVF international conference on computer vision}, 2019, pp. 542--551.

\bibitem{zheng2015partial}
W.-S. Zheng, X.~Li, T.~Xiang, S.~Liao, J.~Lai, and S.~Gong, ``Partial person re-identification,'' in \emph{Proceedings of the IEEE international conference on computer vision}, 2015, pp. 4678--4686.

\bibitem{zhuo2018occluded}
J.~Zhuo, Z.~Chen, J.~Lai, and G.~Wang, ``Occluded person re-identification,'' in \emph{2018 IEEE International Conference on Multimedia and Expo (ICME)}.\hskip 1em plus 0.5em minus 0.4em\relax IEEE, 2018, pp. 1--6.

\bibitem{zhang2019gait}
Z.~Zhang, L.~Tran, X.~Yin, Y.~Atoum, X.~Liu, J.~Wan, and N.~Wang, ``Gait recognition via disentangled representation learning,'' in \emph{Proceedings of the IEEE/CVF Conference on Computer Vision and Pattern Recognition}, 2019, pp. 4710--4719.

\bibitem{zheng2019joint}
Z.~Zheng, X.~Yang, Z.~Yu, L.~Zheng, Y.~Yang, and J.~Kautz, ``Joint discriminative and generative learning for person re-identification,'' in \emph{proceedings of the IEEE/CVF conference on computer vision and pattern recognition}, 2019, pp. 2138--2147.

\bibitem{huang2019celebrities}
Y.~Huang, Q.~Wu, J.~Xu, and Y.~Zhong, ``Celebrities-reid: A benchmark for clothes variation in long-term person re-identification,'' in \emph{2019 International Joint Conference on Neural Networks (IJCNN)}.\hskip 1em plus 0.5em minus 0.4em\relax IEEE, 2019, pp. 1--8.

\bibitem{huang2019beyond}
Y.~Huang, J.~Xu, Q.~Wu, Y.~Zhong, P.~Zhang, and Z.~Zhang, ``Beyond scalar neuron: Adopting vector-neuron capsules for long-term person re-identification,'' \emph{IEEE Transactions on Circuits and Systems for Video Technology}, vol.~30, no.~10, pp. 3459--3471, 2019.

\bibitem{qian2020long}
X.~Qian, W.~Wang, L.~Zhang, F.~Zhu, Y.~Fu, T.~Xiang, Y.-G. Jiang, and X.~Xue, ``Long-term cloth-changing person re-identification,'' in \emph{Proceedings of the Asian Conference on Computer Vision}, 2020.

\bibitem{hong2021fine}
P.~Hong, T.~Wu, A.~Wu, X.~Han, and W.-S. Zheng, ``Fine-grained shape-appearance mutual learning for cloth-changing person re-identification,'' in \emph{Proceedings of the IEEE/CVF conference on computer vision and pattern recognition}, 2021, pp. 10\,513--10\,522.

\bibitem{jin2022cloth}
X.~Jin, T.~He, K.~Zheng, Z.~Yin, X.~Shen, Z.~Huang, R.~Feng, J.~Huang, Z.~Chen, and X.-S. Hua, ``Cloth-changing person re-identification from a single image with gait prediction and regularization,'' in \emph{Proceedings of the IEEE/CVF Conference on Computer Vision and Pattern Recognition}, 2022, pp. 14\,278--14\,287.

\bibitem{fan2020learning}
L.~Fan, T.~Li, R.~Fang, R.~Hristov, Y.~Yuan, and D.~Katabi, ``Learning longterm representations for person re-identification using radio signals,'' in \emph{Proceedings of the IEEE/CVF conference on computer vision and pattern recognition}, 2020, pp. 10\,699--10\,709.

\bibitem{yang2019person}
Q.~Yang, A.~Wu, and W.-S. Zheng, ``Person re-identification by contour sketch under moderate clothing change,'' \emph{IEEE transactions on pattern analysis and machine intelligence}, vol.~43, no.~6, pp. 2029--2046, 2019.

\bibitem{chen2021learning}
J.~Chen, X.~Jiang, F.~Wang, J.~Zhang, F.~Zheng, X.~Sun, and W.-S. Zheng, ``Learning 3d shape feature for texture-insensitive person re-identification,'' in \emph{Proceedings of the IEEE/CVF Conference on Computer Vision and Pattern Recognition}, 2021, pp. 8146--8155.

\bibitem{arkushin2022reface}
D.~Arkushin, B.~Cohen, S.~Peleg, and O.~Fried, ``Reface: Improving clothes-changing re-identification with face features,'' \emph{arXiv preprint arXiv:2211.13807}, 2022.

\bibitem{hermans2017defense}
A.~Hermans, L.~Beyer, and B.~Leibe, ``In defense of the triplet loss for person re-identification,'' \emph{arXiv preprint arXiv:1703.07737}, 2017.

\bibitem{byrne2022fine}
J.~Byrne, G.~Castanon, Z.~Li, and G.~Ettinger, ``Fine-grained activities of people worldwide,'' \emph{arXiv preprint arXiv:2207.05182}, 2022.

\bibitem{wu2019detectron2}
Y.~Wu, A.~Kirillov, F.~Massa, W.-Y. Lo, and R.~Girshick, ``Detectron2,'' \url{https://github.com/facebookresearch/detectron2}, 2019.

\bibitem{liu2023learning}
F.~Liu, M.~Kim, Z.~Gu, A.~Jain, and X.~Liu, ``Learning clothing and pose invariant 3d shape representation for long-term person re-identification,'' in \emph{Proceedings of the IEEE/CVF International Conference on Computer Vision}, 2023, pp. 19\,617--19\,626.

\bibitem{grother2019face}
P.~Grother, M.~Ngan, and K.~Hanaoka, \emph{Face recognition vendor test (fvrt): Part 3, demographic effects}.\hskip 1em plus 0.5em minus 0.4em\relax National Institute of Standards and Technology Gaithersburg, MD, 2019.

\end{thebibliography}
%



%

\begin{IEEEbiography}[{\includegraphics[width=1in,height=1.25in,clip,keepaspectratio]{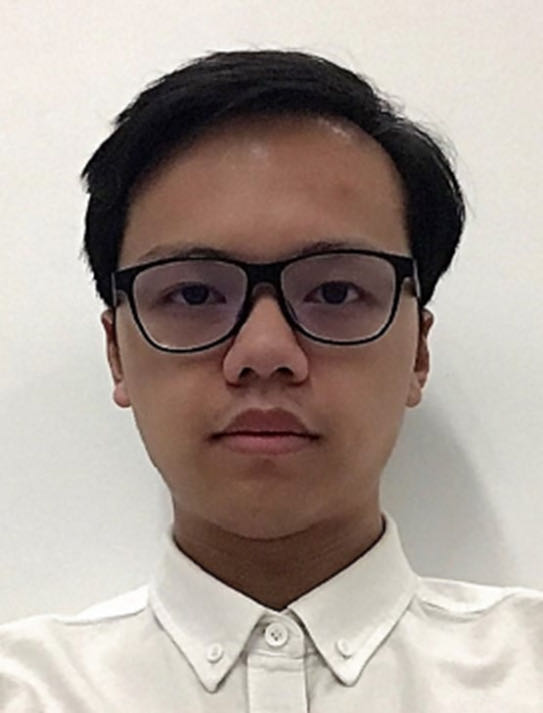}}]{Siyuan Huang}
is a Ph.D. student in Electrical and Computing Engineering at Johns Hopkins University. He was a Graduate Research Assistant at Tsinghua Laboratory of Brain and Intelligence, Tsinghua University, and an Associate Researcher at Alternative Computing Group, National Institute of Standards and Technology. He received a M.S. degree in Computer Science from George Washington University. He works in Prof. Rama Chellappa's lab at JHU on researching person re-identification, body detection, recognition, and identification on large biometric person datasets.
\end{IEEEbiography}

\begin{IEEEbiography}[{\includegraphics[width=1in,height=1.25in,clip,keepaspectratio]{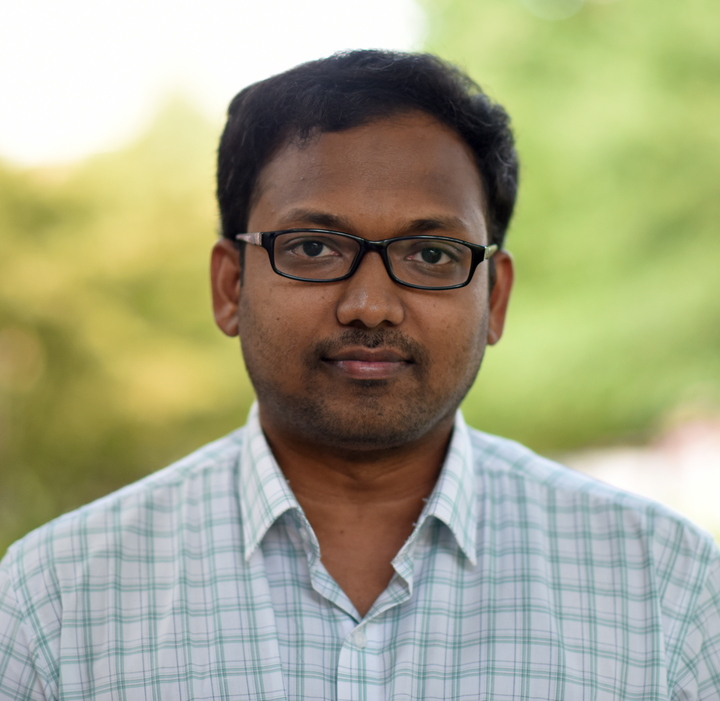}}]{Dr. Ram Prabhakar} is a postdoctoral researcher at Johns Hopkins University. He holds a Master's degree (M.Tech) in Electrical Engineering from National Institute of Technology Rourkela and a Ph.D degree in computer vision from Indian Institute of Science, Bangalore, in 2022. His research interests include computer vision, computational photography, high dynamic range (HDR) imaging, and remote sensing.
\end{IEEEbiography}

\begin{IEEEbiography}[{\includegraphics[width=1in,height=1.25in,clip,keepaspectratio]{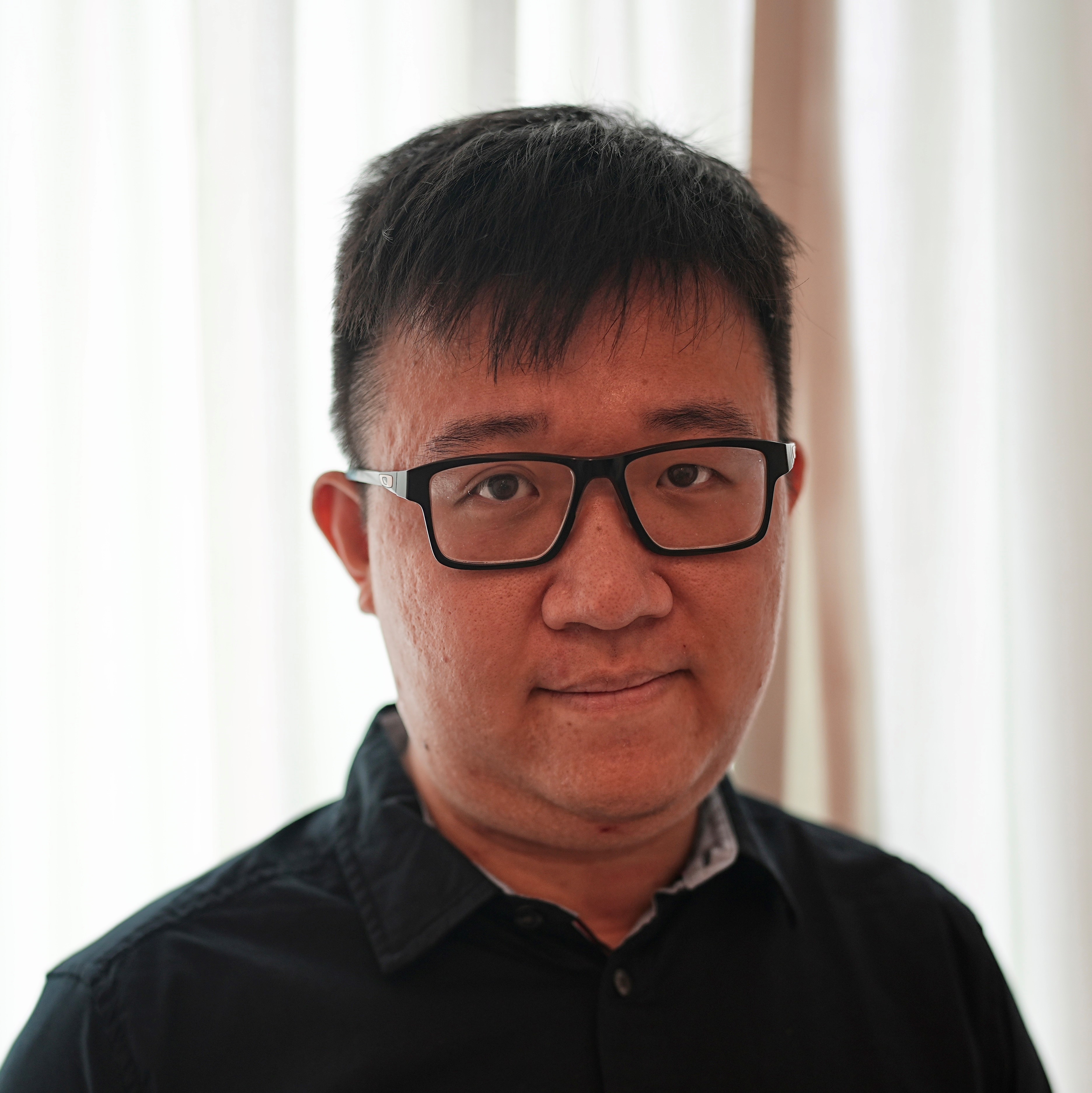}}]{Prof. Chun Pong Lau } is currently an Assistant Professor at the School of Data Science, City University of Hong Kong. He received his Ph.D. in Computer Science in 2021 from Johns Hopkins University. He received his M.Sc. in Applied Mathematics in 2020 from University of Maryland, M.Phil. and B.Sc. in Mathematics from The Chinese University of Hong Kong in 2016 and 2018, respectively. From 2022 to 2023, he was a postdoctoral fellow at Mathematical Institute for Data Science, Johns Hopkins University. With a strong background in mathematics and computer science, Prof. Lau has strong research interests in developing reliable and robust computer vision algorithms, including Atmospheric Turbulence Mitigation, Adversarial Robustness, Biometrics at Severe Conditions, and Generative AI.
\end{IEEEbiography}


\begin{IEEEbiography}[{\includegraphics[width=1in,height=1.25in,clip,keepaspectratio]{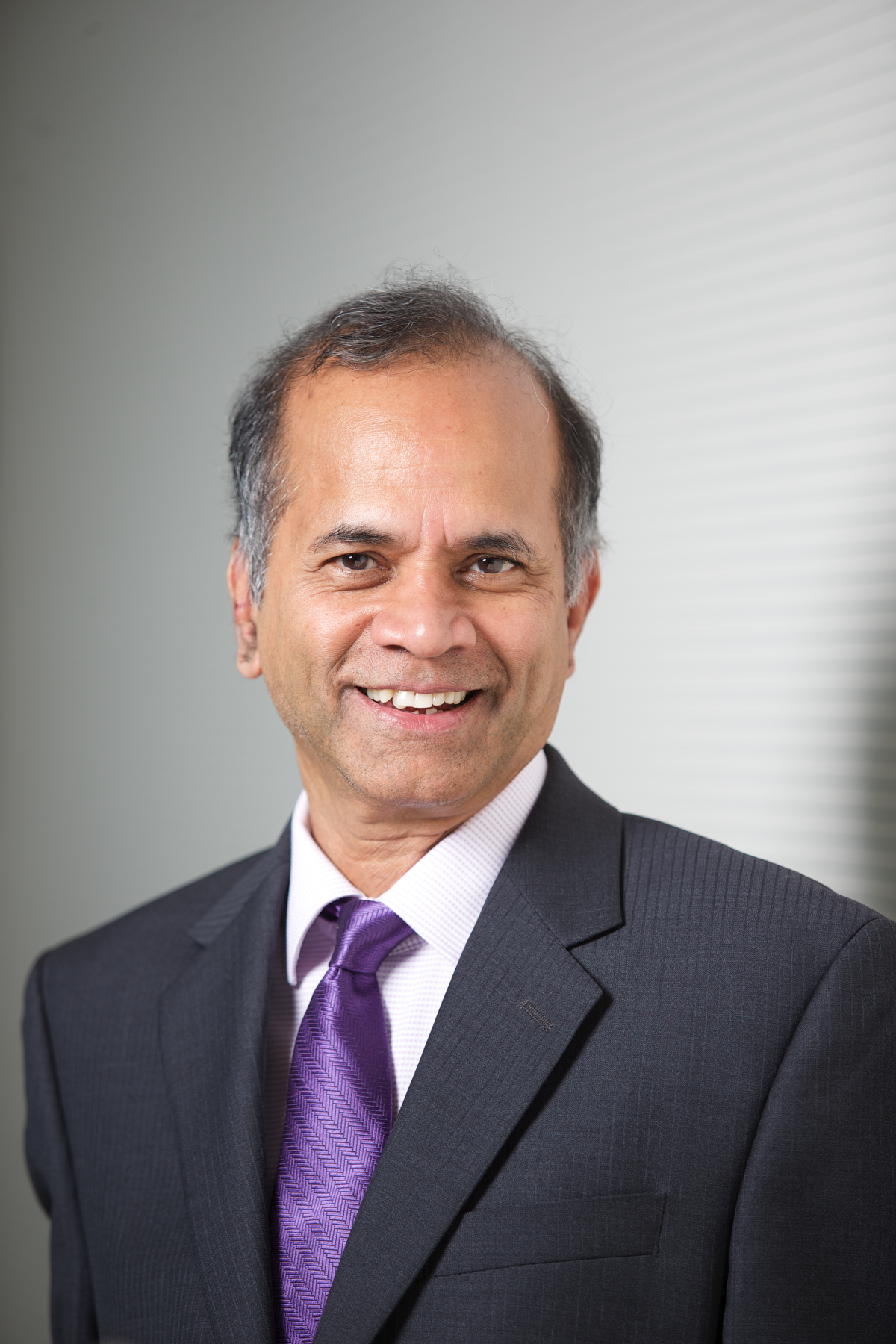}}]{Prof. Rama Chellappa}
is a Bloomberg Distinguished Professor in the Departments of Electrical and Computer Engineering and Biomedical Engineering at Johns Hopkins University (JHU). At JHU, he is also affiliated with CIS, CLSP, IAA, and MINDS. He holds a non-tenured position as a College Park Professor in the ECE department at UMD. His researcher interests are in computer vision, pattern recognition, machine learning and artificial intelligence. He received the 2012 K. S. Fu Prize from the International Association of Pattern Recognition (IAPR). He is a recipient of the Society, Technical Achievement, and Meritorious Service Awards from the IEEE Signal Processing Society, the Technical Achievement and Meritorious Service Awards from the IEEE Computer Society and the Inaugural Leadership Award from the IEEE Biometrics Council. He received the 2020 IEEE Jack S. Kilby Medal for Signal Processing. He is an elected member of the National Academy of Engineering.   He is a Fellow of AAAI, AAAS, ACM, AIMBE, IAPR, IEEE, NAI, OSA, and the Washington Academy of Sciences and holds nine patents.
\end{IEEEbiography}




\end{document}